# Cross-Layer Design for AI Acceleration with Non-Coherent Optical Computing

Febin Sunny, Mahdi Nikdast, Sudeep Pasricha
Department of Electrical and Computer Engineering,
Colorado State University, Fort Collins, CO,
{febin.sunny, mahdi.nikdast, sudeep}@colostate.edu

**ABSTRACT**
Emerging AI applications such as ChatGPT, graph convolutional networks, and other deep neural networks require massive computational resources for training and inference. Contemporary computing platforms such as CPUs, GPUs, and TPUs are struggling to keep up with the demands of these AI applications. Non-coherent optical computing represents a promising approach for light-speed acceleration of AI workloads. In this paper, we show how cross-layer design can overcome challenges in non-coherent optical computing platforms. We describe approaches for optical device engineering, tuning circuit enhancements, and architectural innovations to adapt optical computing to a variety of AI workloads. We also discuss techniques for hardware/software co-design that can intelligently map and adapt AI software to improve its performance on non-coherent optical computing platforms.

## 1. INTRODUCTION

Artificial intelligence (AI) has become an omnipresent technology in recent years, with applications ranging from image processing [1] and indoor localization [2] to autonomous driving [3], [4] and network security [5]. Among various AI techniques being employed, none have been as successful and ubiquitous to machine intelligence as deep neural networks (DNNs). To cater to diverse AI applications, different variations of DNNs have emerged, including convolution neural networks (CNNs), recurrent neural networks (RNNs), graph neural networks (GNNs), and transformers. The common feature across these models is their continuously increasing computational intensity and memory footprint. As such, deploying these DNNs require significant processing power and memory, making it difficult to deploy them on real-world hardware with limited resources.

To meet the growing demands of AI applications, the hardware architecture must scale up in terms of processing capabilities, memory capacity, and communication capabilities. Graphics processing units (GPUs) are usually tasked with accelerating DNN applications todays, but several limitations of the GPU architecture have become apparent in recent years. These limitations include high power consumption and area costs, low performance per watt, and memory bandwidth limitations [6].

The limitations GPU architectures face highlights the need for domain- and application-specific architectures. To address this problem, researchers have been developing various AI accelerator architectures that can provide higher throughput and energy efficiency for AI applications [7]. These accelerators are specialized hardware platforms designed to accelerate the execution of AI algorithms by providing dedicated hardware support for large-scale matrix multiplication, convolution and other common AI operations.

However, the efficiency benefits that can be extracted from electronic AI accelerators are reaching their limits, as the size of transistors approaches the atomic scale. This reduction in transistor size does not translate into reduction in energy consumption as the voltage necessary to drive these transistors has not scaled down proportionally. This leads to increased power density, reduced energy efficiency, and a reduction in expected performance gains. This diminished performance in electronic systems is further exacerbated by data movement bottlenecks created by slow metallic chip-scale interconnects [9].

These disadvantages with electronic accelerators have led to a search for alternative computing technologies that can provide higher performance and lower power consumption. One very promising technology is silicon photonics, which involves the use of light to transmit and process data. Optical computing has the potential to revolutionize AI by providing ultra-fast data transfers and processing while consuming much less energy than traditional electronic computing [12]. Optical computing systems are analog in nature and can perform complex operations, such as multiply and accumulate (MAC) or Fast Fourier Transform (FFT), with low energy consumption and very high speeds.

There are many open challenges with realizing energy-efficient and high throughput optical computing. These challenges can arise from different layers of the system design stack. For a hardware accelerator, the layers of the design stack include the device layer, circuit layer, architecture layer, and software layer. Various noise sources that are difficult to filter out in analog environments, device-level robustness and reliability, speed of operation or lack thereof of auxiliary systems, all pose challenges. To address these challenges effectively, a cross-layer design approach is essential. The cross-layer design process can consider the interactions and dependencies between layers and optimize them jointly to enhance system performance more efficiently than single-layer solutions.

In this paper, we describe challenges with silicon photonic AI acceleration, with a specific focus on the promising non-coherent variant of photonic AI accelerators. We then discuss approaches to overcome these challenges and present case-studies of cross-layer optimized non-coherent photonic AI accelerators.

## 2. PHOTONICS FOR AI ACCELERATION

Silicon photonics offers a promising solution to overcome chip-scale communication bottlenecks faced by electronic (metal) interconnects. Photonic interconnects can realize high-bandwidth, low-latency, and energy-efficient communication that has been shown to outperform traditional metallic interconnects [8]. CMOS-compatible optical interconnects are gradually replacing metallic interconnects, for enabling light-speed communication at almost every level of communication, including at the chip-scale [9]. The same devices that can enable this high-speed communication can be repurposed to perform high-speed computation, such as matrix-vector multiplication, in the optical domain [10].

Photonic AI accelerators can be broadly classified into coherent and non-coherent [12]. Coherent photonic AI accelerators typically use a single wavelength and Mach-Zehnder interferometers (MZIs) devices to manipulate electrical-field phase of an optical signal. The optical phase shift is utilized to perform parameter imprinting and multiplication. Some of the earliest work on photonic AI acceleration was based on coherent architectures [11], where cascaded MZIs arranged in a mesh were used to perform large-scale and high-throughput matrix-matrix and matrix-vector multiplication. There is also increased interest from industry in using coherent architectures for reducing the computational complexity of DNN acceleration at the server scale [12]. However, to achieve reduction in computational complexity, the DNN model must be deployed in its entirety, leading to large MZI mesh sizes and other disadvantages, such as large-area requirement and increased error propagation.

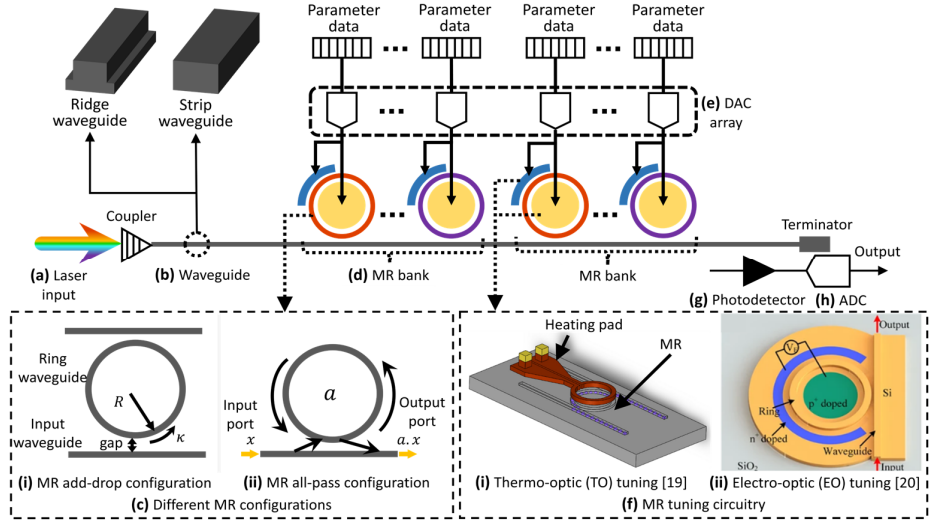

**Fig. 1:** Overview of non-coherent photonic circuit to implement broadcast-and-weight-based multiplication and accumulation for AI acceleration. This circuit needs (a) a laser source, which can be off-chip or on-chip; (b) a waveguide, which can be strip (for passive devices) or ridge (for active devices); (c) microring resonators (MRs), here we show the two popular configurations for MRs, even though (ii) all-pass configuration is used in non-coherent circuits; (d) MR banks perform MAC operation; (e) the banks are tuned as per data from the electronic domain, using digital-to-analog converters (DACs); (f) the tuning circuitry can be conventionally (i) TO or (ii) EO; (g) the result from the MAC operation is present in the collective intensity of wavelengths in the waveguide and can be consolidated using a photodetector; (h) for converting the data to digital domain, for post-processing or storage, an analog-to-digital converter (ADC) can be used.

Non-coherent (a.k.a. in-coherent) photonic AI accelerators make use of multiple wavelengths simultaneously to perform multiple matrix-vector operations concurrently. To imprint parameters, these accelerators typically rely on wavelength-selective filtering and modulation capabilities of microring resonators (MRs). MRs are used to perform amplitude modulation on different wavelengths using a tuning circuit that modifies the operational characteristics of the MRs, to imprint the parameters. Two parameters that need to be multiplied can be deployed on MRs with the same resonant wavelength ($\lambda_{MR}$), along the same input waveguide. The first MR modulates the signal to represent the first parameter. The parameter specific amplitude modulation with the second MR results in the multiplication operation. This approach to multiplication was first conceived as an approach for spike processing in optical spiking neural networks, and is called broadcast and weight [13]. This technique has since been adapted for DNN-specific matrix multiplication operations in various works [14]-[18]. Using the broadcast-and-weight protocol, non-coherent AI accelerator architectures can realize vector-based operations, which can be combined with careful dataflow orchestration to achieve matrix-level operations. To implement such an accelerator substrate, various devices and circuits are required. These photonic devices and circuits are discussed in the following section.

## 3. PHOTONIC DEVICES AND CIRCUITS

A general overview of the fundamental photonic circuit to perform multiplication and accumulation in non-coherent AI accelerators is shown in Fig. 1. In this section, we discuss various devices required for this multiply and accumulate circuit.

*Lasers* are required to generate optical signals necessary for computation and communication in optical circuits. These can be off-chip or on-chip lasers. Off-chip lasers provide better light emission efficiency, but high losses are incurred when the optical signals generated off-chip are coupled to on-chip waveguides. On-chip lasers, such as vertical cavity surface emission lasers (VCSELs), offer better integration density and lower losses. These can be directly modulated lasers which can directly generate modulated signals, or a VCSEL array can provide unmodulated signals which can be then modulated using MZIs or MRs.

*Silicon photonic waveguides* are made of a core (Si) and a cladding ($SiO_2$) material, chosen for high-refractive-index contrast, allowing for total internal reflection and hence optical signal confinement and transmission. These waveguides can be of two types: ridge or strip. A single waveguide can support multiple wavelengths simultaneously without any interference. This is referred to as wavelength-division multiplexing (WDM) and allows for ultra-high bandwidth signal transmission.

*Microring Resonators (MRs)* are made by fabricating a ring-shaped waveguide, in proximity to an input and drop waveguide. This configuration is referred to as the add-drop configuration (see Fig. 1(c)(i)). An MR add-drop filter without the drop waveguide is called an all-pass filter ((see Fig. 1(c)(ii)). MRs can be designed to be sensitive to a particular wavelength, referred to as the MR resonant wavelength $\lambda_{MR}$. The all-pass filter configuration is most commonly used for computation applications. MR banks are groups of MRs sharing a single input waveguide which can be used to perform operations such as multiply and accumulate (MAC) and summation operations in non-coherent architectures.

There are several MR characteristics that are important for the design of non-coherent AI accelerators. The amount of optical signal coupled from the input waveguide to the ring waveguide in an MR is determined by the crossover coupling coefficient, $\kappa$. Q-factor or quality factor of the MR indicates the sharpness of its resonance, or how selective the MR is to its $\lambda_{MR}$. The higher the Q-factor, the more specific the MR wavelength selection and the sharper the response curve will be. Free spectral range (FSR) indicates the spectral distance between two consecutive resonance peaks of an MR. Channel spacing (CS) is the difference between the $\lambda_{MR}$ of two adjacent MR devices along the same input waveguide. Higher CS allows for better isolation between the $\lambda_{MR}$ in the presence of noise, but it limits the number of MRs with different $\lambda_{MR}$ possible along the same input waveguide.

*Tuning circuits* are mechanisms intended to manipulate the effective index ($n_{eff}$) of MR devices, to precisely alter an output optical signal. MRs can leverage electro-optic (EO) or thermo-optic (TO) tuning mechanisms. The TO mechanism (see Fig. 1(f)(i)) uses resistive heat pads over a waveguide to heat up the waveguide, causing changes in $n_{eff}$. The EO mechanism (see Fig. 1(f)(ii)) makes use of doped sections of a waveguide along with a biasing circuit to perform carrier injection, again causing changes in $n_{eff}$.

*Photodetectors (PDs)* are used to detect processed optical signals and convert them to electrical signals. An efficient PD should generate the desired electrical output with a small input optical signal. This input signal power should be larger than the responsivity of the PD. To deliver the appropriate optical power, the optical link design should consider various losses across a link.

## 4. CHALLENGES IN NON-COHERENT PHOTONIC AI ACCELERATORS

Non-coherent silicon photonics for AI acceleration offers a unique solution to the challenges that domain-specific electronic architectures face. But to utilize the advantages provided by these photonic architectures, several challenges must be addressed. In this section, we review some of the challenges in non-coherent photonic AI acceleration, and cross-layer solutions for these challenges that span the device, circuit, architecture, and software layers.

### 4.1. Fabrication-process variations

Fabrication-process variations (FPVs) create unintended changes in fabricated hardware, including both electronic and photonic components, which lead to errors and performance degradation. Thus, it is critical to overcome the impact of FPVs.

The amplitude modulation performed using MRs in non-coherent architectures depend on the wavelength selectivity of the MRs and their resonance-based operation. When an MR is in resonance, the input signal is coupled to the MR with minimal, ideally no, losses. The $\lambda_{MR}$ of an MR is decided by various MR design parameters, including its radius and width and thickness of the waveguides (see Fig. 1(c)(i)). Hence, any change in the MR geometry (e.g., radius, waveguide width and thickness) can cause a resonant shift in the MR ($\Delta\lambda_{MR}$).

FPVs are the primary reason for undesired changes in MR geometry, which can lead to significant errors. In the worst-case, the $\Delta\lambda_{MR}$ can be large enough that modulation intended for one wavelength will be applied to another wavelength. Further exacerbating this challenge is the fact that FPV is more or less random in silicon photonics and can vary from device to device and chip to chip. How FPV will present itself on a chip can be estimated from prior fabrication data, but cannot be predicted with certainty.

In our prior works [21], [22], we addressed FPV for MRs via design exploration and optimization. We showcased a detailed design-space exploration for MRs while optimizing for Q-factor, $\kappa$, and high resilience to FPV. We explored the design space of MRs where input ($w_i$) and ring ($w_r$) waveguide widths are different (see Fig. 2). Such unconventional MR designs explored had $w_i < w_r$, and were shown to have increased FPV resilience, and enhanced $\kappa$ (MR3 design in Fig. 2). The optimal design obtained from the exploration was fabricated and the FPV resilience, in terms of channel-spacing variation ($\Delta c_s$), was verified. These designs were able to achieve 70% reduction in $\Delta\lambda_{MR}$, when compared to conventional MR designs (see Table 1).

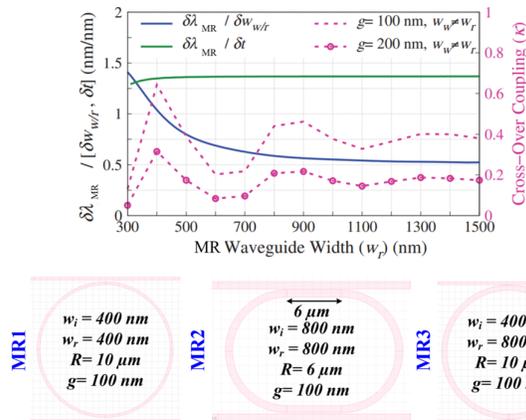

**Fig. 2:** FPV resilient active MRs designed (MR3) as per the technique from [21], showing reduction in $\Delta\lambda_{MR}$ and $\kappa$ retention with increase in $w_r$ (*top*) and the different MR designs that were considered (*bottom*) [22].

**Table 1: Measurement results across MR designs [22]**

| Parameter | MR1 | MR2 | MR3 |
| --- | --- | --- | --- |
| Average $\Delta\lambda_{MR}$ | 7.1 nm | 1.8 nm | 2.1 nm |
| Standard deviation of $\Delta\lambda_{MR}$ | 0.38 nm | 0.17 nm | 0.25 nm |
| Q-factor | 500 | 1800 | 600 |

### 4.2. Thermal crosstalk

Thermal crosstalk impacts the stability of photonic components as well as energy consumption in non-coherent AI accelerators. TO tuning offers large tuning ranges and can be considered for correcting phase perturbations due to FPV or time-varying phase changes due to thermal fluctuations from the electric circuitry in proximity to photonic components. However, TO tuning circuits are susceptible to mutual thermal crosstalk, which reduces MR stability and increases energy consumption.

One solution to this problem would be thermal isolation, i.e., placing MRs sufficiently far apart, so that the heat generated from the resistors are localized. While this may be an acceptable solution in scenarios where area is not an issue, in an accelerator substrate where many thousands of MRs are to be integrated, thermal isolation leads to a significant increase in area consumption. In [14], we built on the Thermal Eigenmode Decomposition (TED) approach from [23] to analyze thermal phase coupling between heating circuits, and apply a phase shift counteracting the phase coupling to the heaters to negate the thermal crosstalk. We used this TED-based tuning to eliminate thermal crosstalk in MR banks for non-coherent optical computing. This approach allowed us to achieve thermal noise-free operation, along with reduction in the overall tuning power consumption. A layout exploration for the MR bank was also conducted to facilitate additional thermal isolation while conserving area. The TED-based exploration in [14] showed a 98.2% reduction in thermal crosstalk noise, along with 68% reduction in tuning power required for MR banks.

### 4.3. Tuning regulated latency of operation

Low latency of operations is crucial to realize high-performance non-coherent photonic AI accelerators. Tuning mechanisms play a key role in determining operation latency as they are essential to achieve changes in parameters being represented by the MRs. TO tuning offers large range of tuning, i.e., large range of $\lambda_{MR}$ changes ($\Delta\lambda_{MR}$). This large range of correction makes TO tuning ideal for FPV correction. But the heating process is slow and within $\mu s$ range, and can also give rise to thermal crosstalk among heaters (Section 4.2). EO tuning is significantly faster and can allow changes in parameters being represented in $ns$ range. But it induces higher losses and cannot offer as high $\Delta\lambda_{MR}$ as TO tuning.

To ensure that the architecture does not incur latency in the $\mu s$ range every time the weights and activations are to be modified, an efficient solution is needed. Given that weight and activation modifications can occur up to thousands of times during a forward pass of a deployed DNN, an efficient tuning approach is crucial in ensuring high throughput and energy efficiency from the accelerator. In [14], we proposed a hybrid tuning approach that utilizes TED to eliminate thermal crosstalk, uses TO tuning to counteract FPV, and then uses EO tuning, due to its low latency, to represent parameters. This allowed the work in [14] to achieve remarkable throughput, up to 15.9× higher when compared to other state-of-the-art AI accelerator platforms.

### 4.4. Signal integrity and crosstalk

The matrix-vector operations performed in the non-coherent photonic AI accelerator require high signal integrity to generate accurate results. In addition to thermal crosstalk (Section 4.2) other sources of signal crosstalk that reduce signal integrity in non-coherent accelerator architectures are heterodyne (incoherent) and homodyne (coherent) optical crosstalk. These are both significant in non-coherent architectures which make use of multiple wavelengths simultaneously via WDM, and may have several devices sensitive to the same $\lambda_{MR}$ along the same input waveguide.

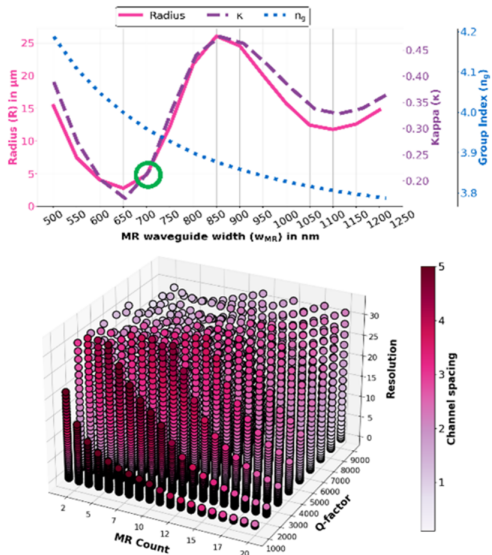

**Fig. 3: Resolution-aware MR design (*top*) and MR-bank design exploration (*bottom*) as proposed in [18].**

Several efforts have analyzed signal crosstalk in photonic circuits [24] as signal crosstalk impacts achievable resolution for parameters. Here resolution refers to the number of bits used to define the parameter value. In digital environments, such as CPUs and GPUs, parameter resolution is usually at 32-bits. In an analog setting, to represent an n-bit parameter, the device should be able to represent $2^n$ levels. These levels are represented as different signal amplitude levels in non-coherent photonic signals. The presence of noise can lead to errors as the levels can get misinterpreted at the output. At higher parameter resolution, signal crosstalk can impact the model accuracy significantly [24] and hence measures must be taken to mitigate signal crosstalk.

To facilitate high parameter resolution, we proposed to use MR designs which had a Q-factor of 8000, allowing for better spectral isolation between neighboring MRs [14]. Because heterodyne crosstalk arises from overlap in frequency response of neighboring MRs, spectrally isolating them mitigates this type of crosstalk. The resolution was modeled with an inverse relationship to the overall heterodyne noise presence in the MR bank. Using this heterodyne crosstalk model along with the MR designs, we achieved a high resolution of 16-bits in [14]. Our work in [18] followed a similar approach to achieving 16-bit resolution, but made device design considerations to avoid higher order mode excitation, and optical losses in MRs. The resulting MRs had a ring waveguide width of 700 nm, radius of 5$\mu m$, and a Q-factor of 5000. To improve the spectral isolation, this work considered increasing channel spacing, with a channel spacing of 2.5 nm between adjacent $\lambda_{MR}$ (see Fig. 3).

### 4.5. Electro-optic conversions

The high power/energy consumption at electro-optic interfaces is a major challenge in non-coherent photonic AI accelerators. To drive tuning circuitry, we need to convert digital representations of the parameters into the analog domain and to convert results from the photonic domain back to the digital electronic domain for post processing and storage. For these conversions, we need to use analog-to-digital converters (ADCs) and digital-to-analog converters (DACs), respectively. The number of required DACs exceeds the number of ADCs needed, by a large margin. In Section 4.4, we discussed how parameter resolution is important, but what we did not discuss there was the cost of ensuring high resolution of operation. A high parameter resolution dictates that the DACs and ADCs that are used in the architecture should also be of the same resolution or higher. It should be noted that at higher resolutions, DAC and ADC power increases exponentially.

In an attempt to limit the DAC and ADC power consumption, while sustaining model accuracy, in [14] we quantized DNN models to 16-bit resolution (from 32-bit resolution) in the digital domain where they were trained. But this 16-bit resolution still comes at the cost of high power consumption and latency, mainly due to the DACs. One of the most effective ways to handle this, without incurring errors in programming the parameters into the photonic substrate, is to further quantize the DNN model being deployed directly. AI models can be quantized to much smaller parameter sizes, below 16-bit quantization. An extreme example of this is binarized neural networks or BNNs [25].

BNNs use binary values for weights and activation values, which degrades their output accuracy, but helps them achieve significantly smaller memory footprint. This makes them ideal (for some applications) to be deployed in resource-constrained edge and IoT environments. Our work in [15] focused on how to accelerate BNNs, and showcased how increasing activation parameter size along with batch normalization can improve top-5 accuracies of the models being deployed. Our analyses showed that at 4-bit activation size, the accuracy benefits diminish for the BNN models and hence the architecture was designed to support 1-bit weights and 4-bit activations. Using these partial BNNs only requires simple switching circuits for weight parameter representation and simpler DAC circuitry for activations, leading to better performance in terms of throughput and energy consumption. Having less stringent constraints on number of amplitude levels necessary for parameter representation makes BNNs more resilient to the noises in the analog domain as well. From an analysis on BNN's resiliency to these noises, [15] showed that a further 20% of tuning power can be saved in the proposed architecture (see Fig. 4).

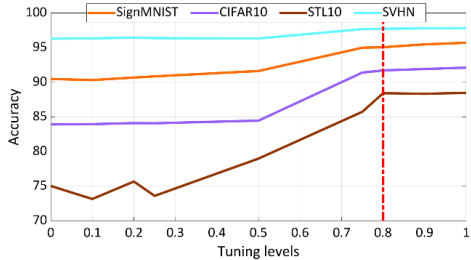

Fig. 4: Inference accuracy for different models vs. level of tuning as explored in [15]. At 80% tuning, accuracy saturates, making further tuning unnecessary, and providing an opportunity to save tuning power.

### 4.6. Efficient matrix computations

The manner in which DNN computations are decomposed and orchestrated is important to achieve high performance and energy efficiency in non-coherent AI accelerators. In DNN models, the matrix operations can be decomposed into vector-dot-product operations. Our work in [14]-[18] focused on implementing high throughput and energy-efficient vector-dot-product units (VDUs) using non-coherent photonics. As an example, to achieve efficient matrix computation, in [14], the photonic accelerator was divided into two main sections, with each focused on accelerating one specific layer type of CNNs (fully connected or convolution). Accordingly, The VDU types were designed to accelerate convolution layers and fully connected layers. The vector granularity of the VDUs, i.e., maximum size of vector that can be deployed to them, was also optimized via design-space exploration.

Table 2: Average EPB and FPS/Watt values across accelerators [14].

| Accelerator | Avg. EPB (pJ/bit) | Avg. kiloFPS/watt |
|---|---|---|
| P100 | 971.31 | 24.9 |
| IXP 9282 | 5099.68 | 2.39 |
| AMD-TR | 5831.18 | 2.09 |
| DaDianNao | 58.33 | 0.65 |
| Edge TPU | 697.37 | 17.53 |
| Null Hop | 2727.43 | 4.48 |
| DEAP_CNN | 44453.88 | 0.07 |
| Holylight | 274.13 | 3.3 |
| Cross_base | 142.35 | 10.78 |
| Cross_base_TED | 92.64 | 16.54 |
| Cross_opt | 75.58 | 20.25 |
| Cross_opt_TED | 28.78 | 52.59 |

## 5. ANALYSIS OF CROSS-LAYER SOLUTIONS

To showcase the effectiveness of combining cross-layer design approaches, we describe results of cross-layer optimization across various non-coherent photonic AI accelerators [14]-[18].

First, we present a sensitivity analysis from our work in [14]. The baseline non-coherent photonic AI accelerator architecture was simulated with no optimization other than VDU design (*Cross_base*), and contrasted with the baseline architecture with TED (*Cross_base_TED*), with optimized FPV resilient MRs (*Cross_opt*), and finally the fully cross-layer optimized architecture (*Cross_opt_TED*). From this analysis, it was shown that the fully cross-layer optimized design is able to achieve up to 9.5× lower energy-per-bit (EPB) of operation than other state-of-the-art accelerators of the time (see Table 2). In the table, P100 is Nvidia P100 GPU, IXP 9282 is Intel Xeon Platinum 9282 CPU, and AMD-TR is AMD Threadripper CPU. DaDianNao [26], EdgeTPU, and NullHop [27] are electronic accelerators, while DEAP-CNN [28] and HolyLight [29] are photonic accelerators. FPS in the table stands for frames-per-second and is a measure of throughput. FPS/Watt is thus a measure of throughput efficiency.

The VDUs designed to accelerate BNNs in [15] followed a similar floorplan as the one in [14]. However, as the parameter resolution requirements were very different, the device-level requirements also change. BNNs also make use of batch normalization (BN) to boost their accuracies, which has to be included in the forward pass of the inference stage as well. To address these requirements, the VDUs in [15], depicted in Fig. 5, make use of three separate MR designs. A 1.5 $\mu m$ radius conventional MR is used for the binary weight value representation. This reduced MR size helped reduce waveguide lengths and associated losses. The 4-bit activations were represented using 5 $\mu m$ non-conventional MRs with 760 nm ring waveguide width for increased FPV tolerance. The proposed VDU also used a broadband MR design for implementing the BN parameter. The broadband MR facilitates a multiplication operation on the multiple output wavelengths simultaneously, without the need for an optical-to-electrical conversion and storage.

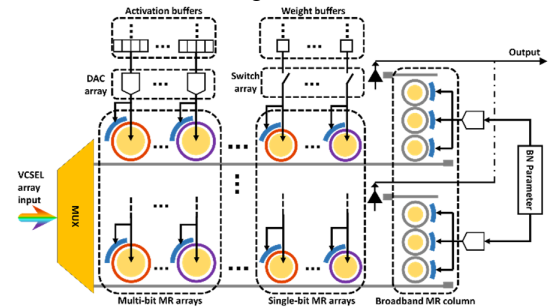

Fig. 5. BNN-specific VDU design from [15].

BNNs offer reduction in memory footprint, but at the cost of increased errors. DNN models can still achieve reduction in memory footprint, but retain or even improve on output accuracy through sparsity exploitation. Sparsity in a DNN model refers to removal of weights, i.e., replacing them with 0s, which does not contribute to the overall accuracy of the model. But this technique gives rise to another challenge. The sparse models incur unwanted operations which ultimately lead to a 0 output because of the 0 valued weights involved. This leads to wasted energy consumption and latency.

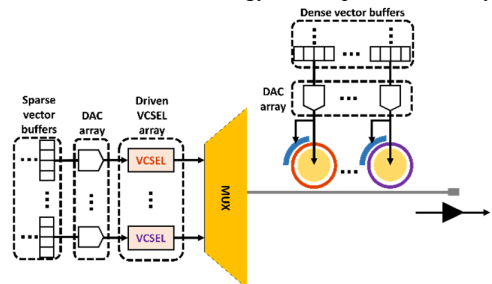

Fig. 6: VDU design for sparse CNN acceleration, from [16].

Our work in [16] tackled this challenge by combining a dataflow mechanism for compressing sparse matrices to dense matrices along with VDU design to handle any residual sparsity. The VDU in this architecture makes use of VCSELs to directly tune the wavelength to reflect the parameter followed by an MR bank to perform the dot-product operation (see Fig. 6). Vector granularity-aware VDU designs were implemented for convolution and fully connected layers as well. This work was able to leverage sparsification of the CNN matrices along with a weight-clustering-based quantization to generate sparse and quantized CNN models for deployment. These models showed similar, or sometimes even better, accuracies than the base models (see Table 3). The software design approach along with the VDU design for [16] helped achieve 27.6× lower EPB (see Fig. 7) than other state-of-the-art accelerators.

Table 3: A summary of the sparsification and quantization results across models from [16].

| Dataset | No. of layers | Layers pruned | Sparsity induced | Quant. level | ΔAccuracy |
|---|---|---|---|---|---|
| MNIST | 4 | 4 | 50% | 6-bit | -0.31% |
| CIFAR10 | 7 | 7 | 47.1% | 4-bit | +0.81% |
| STL10 | 7 | 5 | 40% | 6-bit | +0.6% |
| SVHN | 7 | 5 | 41.1% | 6-bit | +0.4% |

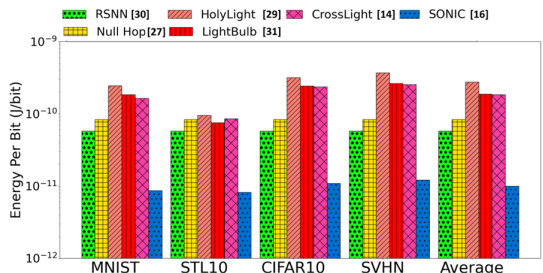
Fig. 7: EPB comparison for accelerators across CNN models [16].

Another approach to reduce memory footprint while retaining baseline accuracy in DNN models is to consider heterogeneous layer-wise quantization. To efficiently accelerate such models, there are several challenges to overcome. Because the parameter resolutions can vary from layer to layer, the hardware cannot have a fixed resolution. Using a large DAC/ADC interface can lead to energy wastage, as the large resolution leads to large power consumption, and this resolution may not be used by all the layers. Accordingly, a bit-slicing-based parameter deployment was adopted in our architecture in [17] called HQNNA. But unlike electronic architectures that use multiple devices to deploy bit-sliced parameters, photonic architectures cannot afford spare devices. This is because photonic devices are bulky and the presence of un-tuned devices can still lead to crosstalk and energy wastage during signal propagation. To address this, [17] adopted a dataflow orchestration where bit slices are deployed across time steps to achieve the necessary operation. This approach led to some reduction in throughput, but achieves up to 73.8× better EPB (see Fig. 8) than conventional photonic accelerators. In [18], we applied many of these cross-layer optimizations to accelerate the family of recurrent neural networks.

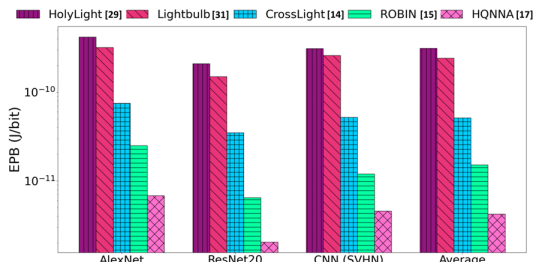
Fig. 8: EPB comparison for CNN accelerators and HQNNA [17].

## 6. CONCLUSIONS

Silicon photonics for AI acceleration represent a paradigm shift in computing, towards high-throughput, energy-efficient AI. Non-coherent photonic computing offers a unique solution which can still leverage the various software-level and dataflow optimization intended for electronic accelerators while providing superior performance benefits with silicon photonics. To realize this promise, cross-layer system design and optimization is the key. In this paper, we have provided an overview of various challenges that non-coherent AI acceleration faces and our solutions to those challenges, through cross-layer design solutions.

## ACKNOWLEDGEMENTS
This work was supported by National Science Foundation (NSF), through grants CCF-1813370 and CCF-2006788.